\documentclass[conference]{IEEEtran}
\IEEEoverridecommandlockouts
\usepackage{cite}
\usepackage{amsmath,amssymb,amsfonts}
\usepackage{algorithmic}
\usepackage{graphicx}
\usepackage{textcomp}
\usepackage{xcolor}
\usepackage{subfigure}
\usepackage{multirow}
\def\BibTeX{{\rm B\kern-.05em{\sc i\kern-.025em b}\kern-.08em
    T\kern-.1667em\lower.7ex\hbox{E}\kern-.125emX}}
\begin{document}

\title{DEPHN: Different Expression Parallel Heterogeneous Network using virtual gradient optimization for Multi-task Learning}

\author{
	\IEEEauthorblockN{
		Menglin Kong\IEEEauthorrefmark{1},
		Ri Su \thanks{ Ri Su is corresponding author.}\IEEEauthorrefmark{2},
		Shaojie Zhao\IEEEauthorrefmark{3},
		Muzhou Hou\IEEEauthorrefmark{4}
		}
	\IEEEauthorblockA{\IEEEauthorrefmark{1}\textit{School of Mathematics and Statistics}\\
\textit{Central South University}, Changsha, China, Email: 212112025@csu.edu.cn}
	\IEEEauthorblockA{\IEEEauthorrefmark{2}\textit{School of Mathematics and Statistics}\\
\textit{Central South University}, Changsha, China, Email: suricsu@csu.edu.cn}
	\IEEEauthorblockA{\IEEEauthorrefmark{3}\textit{School of Mathematics, Physics \& Statistics}\\
\textit{Shanghai University of Engineering Science}, Shanghai, China, Email: m440121303@sues.edu.cn}
	\IEEEauthorblockA{\IEEEauthorrefmark{4}\textit{School of Mathematics and Statistics}\\
\textit{Central South University}, Changsha, China, Email: houmuzhou@sina.com}
}

\maketitle

\begin{abstract}
Recommendation system algorithm based on multi-task learning (MTL) is the major method for Internet operators to understand users and predict their behaviors in the multi-behavior scenario of platform. Task correlation is an important consideration of MTL goals, traditional models use shared-bottom models and gating experts to realize shared representation learning and information differentiation. However, The relationship between real-world tasks is often more complex than existing methods do not handle properly sharing information. In this paper, we propose an Different Expression Parallel Heterogeneous Network (DEPHN) to model multiple tasks simultaneously. DEPHN constructs the experts at the bottom of the model by using different feature interaction methods to improve the generalization ability of the shared information flow. In view of the model’s differentiating ability for different task information flows, DEPHN uses feature explicit mapping and virtual gradient coefficient for expert gating during the training process, and adaptively adjusts the learning intensity of the gated unit by considering the difference of gating values and task correlation. Extensive experiments on artificial and real-world datasets demonstrate that our proposed method can capture task correlation in complex situations and achieve better performance than baseline models\footnote{Accepted in IJCNN2023}.
\end{abstract}


\section{Introduction}\label{1}
In the real-world internet service scenario, user behaviors are usually multifaceted, such as "browsing, clicking, collecting, buying, returning" in the extent of e-commerce, and "click, like, finish broadcast, launch, review" in the streaming media scene\cite{2018End}. Under the same internet application, network service providers analyze various behaviors of users for better understanding of user positioning, to provide users with better personalized recommendation services\cite{2019Deep}. Under the guidance of multi-objectives, the model will also receive more accurate feedback and better analyze the platform content and user needs. In the existing solutions, the end-to-end multi-task model caters to the service needs of internet users. Multi-task learning ({\bf MTL}) aims to use a common model to fit multiple task objectives in the same user scenario simultaneously, so as to make full use of the information flow between users and objects\cite{LVB2021Deep}.

\begin{figure}[ht]
  \centering
  \addtocounter{subfigure}{0}\subfigure[MTL with shared bottom]{
    \includegraphics[width=.13\textwidth]{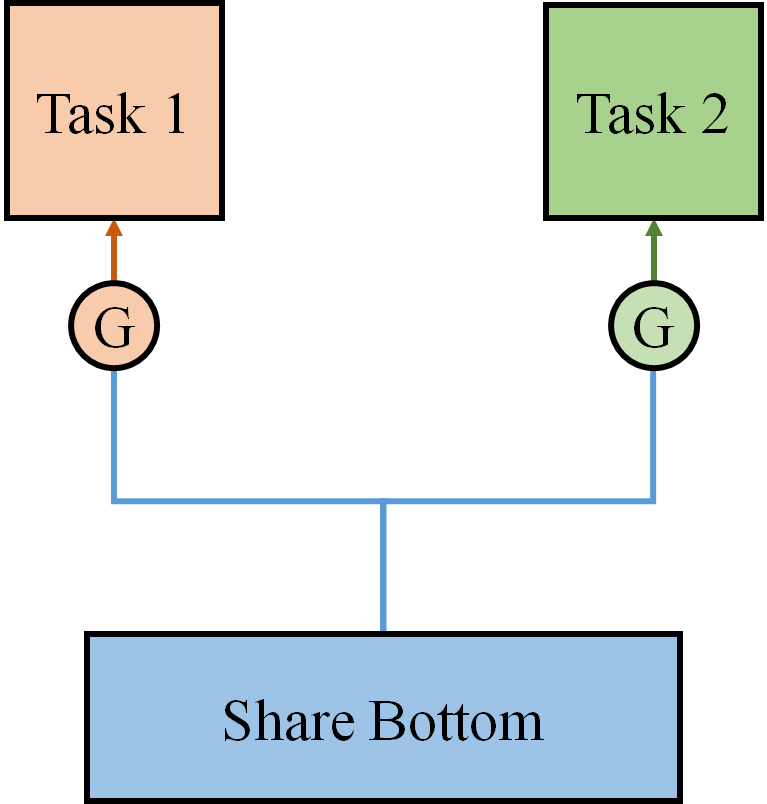}
  }\hspace{.02\textwidth}
  \addtocounter{subfigure}{0}\subfigure[MTL with information transaction]{
    \includegraphics[width=.13\textwidth]{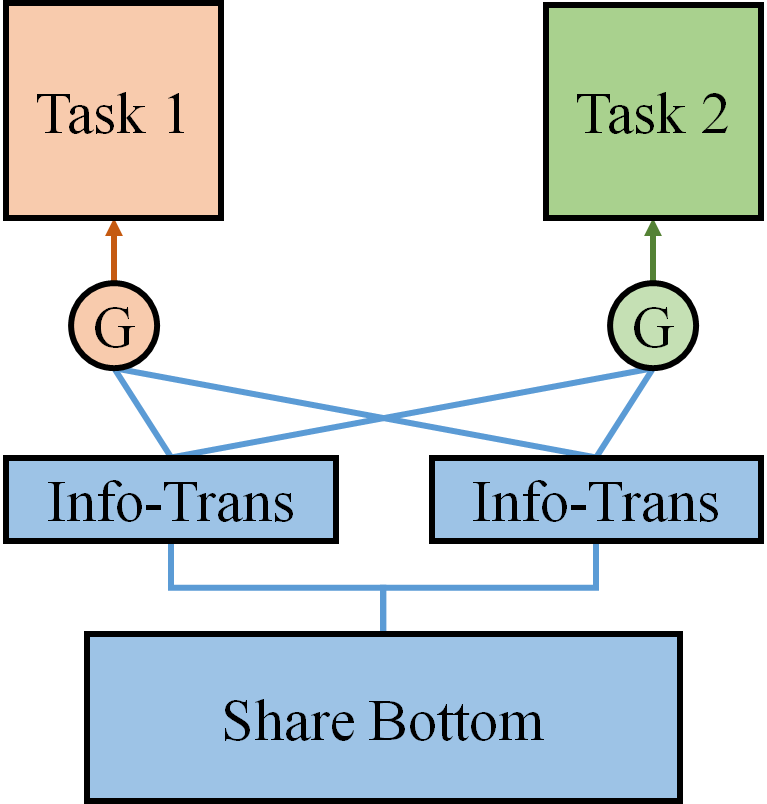}
  }\hspace{.02\textwidth}
  \addtocounter{subfigure}{0}\subfigure[MTL with Task-related involvement]{
    \includegraphics[width=.13\textwidth]{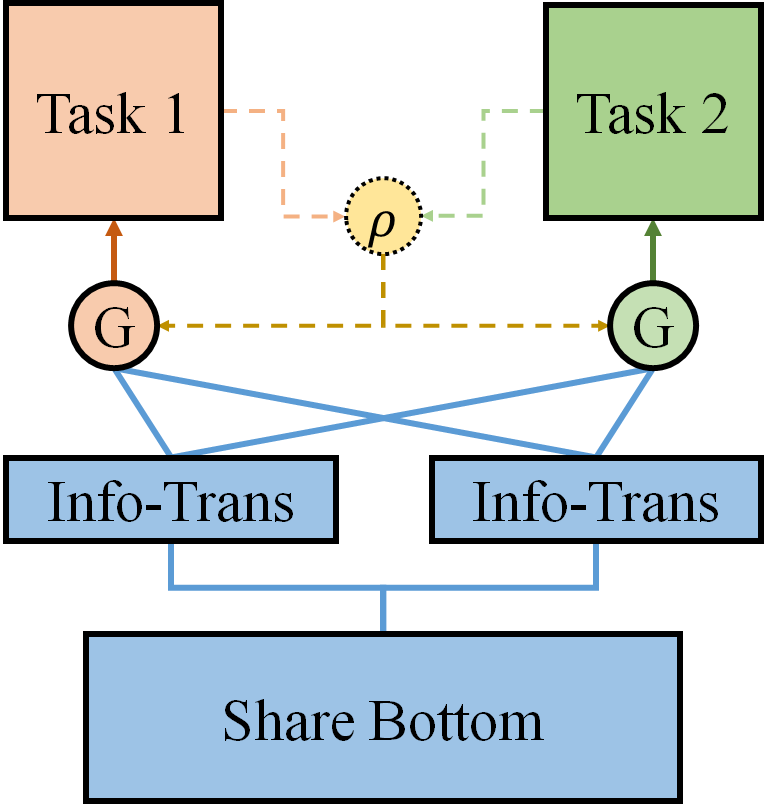}
  }
  \caption{The differences of MTL basic framework in recommendation system.}
  \label{pic:mtl-model}
\end{figure}

In the traditional MTL model, the existing research usually uses the trainable shared bottom layer to enhance the generalization ability of the model, and its basic framework is shown in Figure\ref{pic:mtl-model}(a). In view of the information flow difference between different tasks, how to improve the information differentiation ability of the sharing bottom part has become an important aspect of the multitasking model\cite{2018Multi}. Among the existing deep learning models, MMoE\cite{MMOE}, PLE\cite{2020Progressive} and SNR\cite{2019SNR} are prevalent baselines in the field of MTL, which aim at sharing and differentiating of features. From another perspective, we can view the sharing layers of multiple experts as a whole, representing the various elements of shared information. As shown in Figure\ref{pic:mtl-model}(b), in the back propagation stage of the training process, the task tower uses the returned gradient to find the optimal gating value, so as to obtain the most favorable information distribution for the task.

Task correlation is another important consideration of MTL goals, while the joint learning of multiple tasks with low correlation often requires more differentiation in the sharing of information. Traditional models use shared-bottom models and gating experts to realize shared representation learning and information differentiation, but the real-world significance of tasks often has a more complex relationship. As shown in Figure\ref{pic:mtl-model}(c), incorporating the correlation between tasks into the screening of information is a better approach. AITM\cite{2021Modeling} also points out that in different scenarios, there is more of a sequential dependence relationship between the objectives of multi-task optimization. Therefore, for more complex scenes, controlling the information flow of tasks more accurately is necessary, so as to achieve the purpose of improving the accuracy of all tasks through multi-task joint optimization.

Along this line, we propose an Different Expression Parallel Heterogeneous Network ({\bf DEPHN}) to model multiple tasks simultaneously. Specifically, in order to improve the generalization ability of the shared information flow of the MTL model, DEPHN constructs the experts at the bottom of the sharing system by using different feature interaction ways in the field of recommendation system, and applies the soft gating mechanism ({\bf SSG}) to integrate the information for different types of expert components. Secondly, in view of the model's differentiated ability requirements for different task information flows, DEPHN uses explicit mapping with virtual gradient indicator coefficient for experts output at the training process, and adaptively adjusts the learning intensity of the gated unit by considering the difference of gating values and task correlation.

This paper mainly contributions are as follow:
\begin{itemize}
	\item We use different feature interaction modules in PHN\cite{PHN} and discuss the private components and public components in the shared bottom layer for the specific and shared information of tasks. Using SSG module to strengthen the selection and combination of raw features for different feature crossing modules. 
	\item From the perspective of function decomposition, we carried out explicit mapping of the activation values output by different experts, which strengthens the information flow representation method of the model, and also provides more information integration schemes for the subsequent gated units.
	\item Virtual Gradient is added in the training process, and the correlation measurement between tasks is transmitted back to the corresponding gated unit as a gradient, so that it can focus on the fitting of task similarity and differences and caters to the design of explicit mapping of expert outputs.
	\item We use artificial and real data sets to verify the effectiveness of PHN, explicit mapping and virtual gradient in the multi-tasking model. DEPHN also provides explicable possibilities for understanding the underlying relationships of the model.
\end{itemize}

The rest of this paper in organized as follows: Section 2 reviews related work. Section 3 clarify the basic framework of MTL and give a definition. Section 4 introduces the proposed approach in details. Section 5 evaluates the proposed models. Finally, Section 6 concludes the paper.

\section{RELATED WORK}\label{2}

\subsection{MTL in Recommendation System}\label{2.1}
Many recommendation models can predict multiple tasks at the same time\cite{ESMM}. The classic ESMM model\cite{ESMM} introduces pCTCVR prediction as auxiliary task and to train the pCVR main task tower and the pCTR task tower. The embedding layer of the two towers is shared and the entire model is trained in the full sample space, so as to alleviate the problem of training data sparsity of pCVR. ESCM2 still aims to solve various problems of CVR prediction, and solve the two major problems of biased prediction and independent priori existing in ESMM with the help of causal inference\cite{2022ESCM}. AITM\cite{2021Modeling} puts the research focus on the multi-task modeling with subsequently dependent order. This framework takes into account the influence of the previous task on the later task, weights the information of the previous task and the current information through the attention mechanism, and inputs it to the tower layer of the current task to complete the prediction of the current task.

\subsection{Task relation Study}\label{2.2}
Many researchers have proposed an end-to-end model based on shared bottom layers from the perspective of alleviating the overfitting phenomenon of single task with the help of MTL\cite{2020Progressive} . Among the existing models, Cross-Stitch\cite{2016Cross} modeles two bottom-shared base networks for the two tasks respectively, and introduces the cross unit of sharing parameters to realize the dual transfer of information within the two networks . Sluice\cite{ruder2017sluice} strengthens the training intensity of sharing bottom by using residual connection. MMoE\cite{MMOE} mainly focuses on information screening, using gated units to integrate information of experts at the bottom of the task tower; PLE\cite{2020Progressive} focuses on information sharing and integrating between feature and experts, and further optimizes task-specific information and task-shared information in a multi-level feature extraction manner. SNR\cite{2019SNR} provides a different way of information sharing and introduces random variables as coefficients of model weights to find the optimal model structure and select the way information flowing inside the model.

\section{METHODOLOGY}\label{3}
\begin{figure*}[ht]
  \centering
  \includegraphics[width=0.7\textwidth]{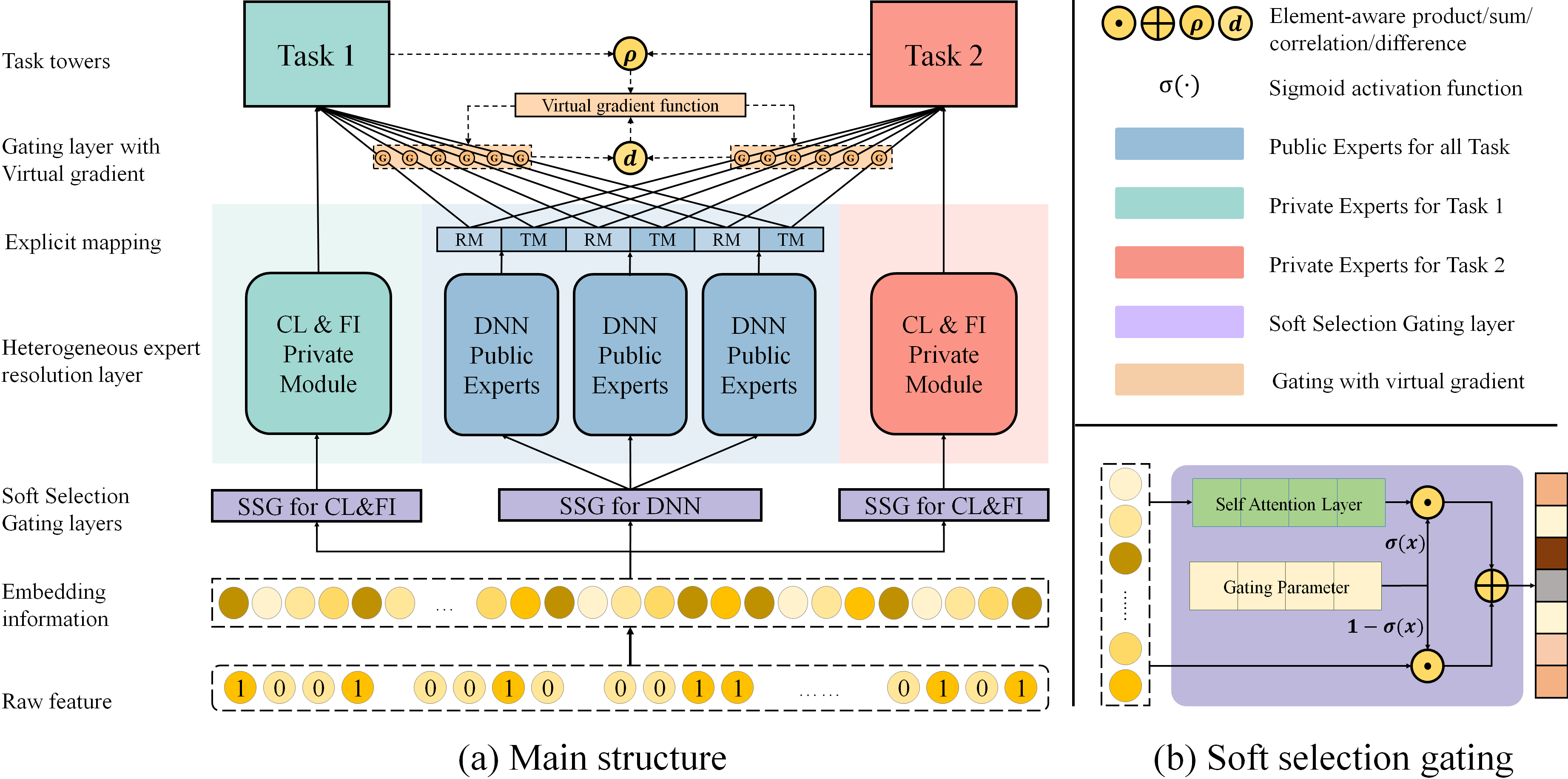}
  \caption{The structure of DEPHN from input to output includes: soft gating layer, expert analysis layer, virtual gradient based gating layer and finally task tower.}\label{pic:stru}
\end{figure*}

This section describes the structure of Different Expression Parallel Heterogeneous Network ({\bf DEPHN}). Figure \ref{pic:stru} shows the main structure of the model.

Firstly, DEPHN uses different feature analysis patterns to explore the private cross patterns and public information characteristics of tasks, and also uses soft gating mechanism ({\bf SSG}) and residual link with trainable parameters to ease the weak gradient phenomenon. Secondly, DEPHN explicitly maps public information to improve the expression dimension of shared information flow. Finally, in order to strengthen the public part of the gate unit for information diversion, the virtual gradient coefficient based on task correlation and information flow difference is used in the training process to improve the representation of the model under low task correlation.

\subsection{Parallel Heterogeneous Layers}\label{3.1}
The Parallel Heterogeneous Layers consists of two main components, Soft Selection Gating({\bf SSG}) module and Heterogeneous Interaction Layer ({\bf HIL}). SSG module based on self-attention to enhance embedding features for different structures, and HIL using different interaction method to analyze the enhanced features.

\subsubsection{Soft Selection Gating (SSG)}\label{3.1.1}
In order to strengthen the information analysis ability of different cross modules, the traditional Multi-head self-attention({\bf MSA})\cite{transformer} used to aggregate strong influence relationships between input information to improve the representation of the information to the predicted distribution. MSA has considered different field weights through the interaction of query vector, key vector and value vector. However, the direct using of the traditional MSA may over-focus on the feature activation value of the low-order interaction, thus losing the information of the feature at the higher-order crossover and raw feature. As the Figure\ref{pic:stru}(b) shows, an information selecting method named Soft Selection Gating (SSG) is used after the sharing embedding $E \in \mathbb{R}^d$ in DEPHN.
\begin{equation}
	E_{sg} = G_{sg}\odot E_{SA}+[I-G_{sg}] \odot E,
	\label{eq:SSG}
\end{equation}
where $G_{sg} \in \mathbb{R}^d$ is the trainable vector, $E_{SA} \in \mathbb{R}^d$ is the MSA embedding $E \in \mathbb{R}^d$ is the raw embedding, and $I \in \mathbb{R}^d$ is the identity vector, $\odot$ is Hadamard product. In order to introduce Shared and exclusive characteristics\cite{2020Progressive} of the information flow on the input side, the SSG considers both sharing the raw embedding and the MSA embedding. By using gating parameters $G_{sg}$, the model can select the raw feature or the feature enhanced by MSA for different parallel structures. Subsequent experiments will discuss whether different structures have different preferences for input features and confirm the effectiveness of SSG.

\subsubsection{Heterogeneous Interaction Layer{HIL}}\label{3.1.2}
The HIL used three kinds of interaction methods to improve interaction capability of the bottom layers: 1) Cross Interaction Layer consists of the feature explicit interaction ways of DCN\cite{dcn} and DCNV2\cite{dcnv2}, which achieved the data mode of high-order interaction by realizing the intersection of multi-layer hidden features and original features.; 2) Field Interaction Layer is mentioned in FINT\cite{FINT}, which uses the cross method to implement the vector-wise interaction. 3) DNN modules is used to fit the non-polynomial information. In these methods, DNN is used as the task-sharing part for its capacity to extract task-independent semantic in formation by implicitly conducting high-order feature interaction, and subsequent experiments will confirm this settings. Based on MMoE, the model in this section is defined as the multi-task parallel heterogeneous network ({\bf MTPHN})

\subsection{Explicit mapping for function decomposition}\label{3.2}
At the beginning of this section, to clarify our approach, we define,
\begin{equation}
	z_{pub} \triangleq \left[h_{DNN}^0(field=1)\left |  \right | \cdots \| h_{DNN}^0(field=c)\right],
	\label{eq:def}
\end{equation}
where $\left |  \right | $ represents concatenation of vectors, $h_{DNN}^0(field=1)$ is the input feature of sharing DNN modules from field 1. Similarly, $z_{pri}$ can be defined based on $h_{CL}^0(field=1)$ or $h_{FI}^0(field=1)$, which are the input features of the private expert cross interaction layer and field interaction layer respectively.

With regard to the goal of MTL, we assume that the correlation between tasks can be represented by the activation values represented under the label distribution. That is, we can assume that under the label representation, there are activation values that describe how interested a user is in an item. The functional relationship presented by the activation values in different tasks is exactly the target that the multi-task model needs to approximate.

For two different tasks $\mathcal{T}_A$ and $\mathcal{T}_B$, we assume that there is a relationship between them potential activation values $a$ and $b$:
\begin{equation}
	b=f_t(a)\approx\sum_{k=1}^{K}f_{k}(a)),
	\label{eq:hypo}
\end{equation}
where  $f_t{\cot}$ is the functional relationship the multi-task model need to approximate. In the setting of multiple experts, the model uses the expression pattern $f_{k}$ of $K$ experts to approximate the target function $f_t{\cot}$. To find an implicit mapping $f_{k}$ of expert k within tag values, inspired by the famous \textbf{Taylor's Theorem} and \textbf{Fourier series} \cite{stewart2015calculus}, explicit mapping operations is carried out on the activation values output by expert k.

From the perspective of function expansion, the explicit mapping module uses nonlinear functions ('RM' as raw mapping and 'TM' as trigonometric mapping in Fig\ref{pic:stru}(a)) to fit the trend relationship and periodic relationship between the activation values of different tasks. It explicitly constructs the activation values output by experts of the common module, so as to strengthen the representation of experts. After the explicit mapping of the same expert in the later stage, the exclusive gating of different tasks will integrate the information flow, screen the required information and send it to the final task tower.

The overall feedforward process of DEPHN is as follows,
\begin{equation}
	\widehat{y_{t}^{pub}}=\sigma(\sum_{k=1}^{K_{p u b}} \sum_{p=1}^{P} W_{k p}^{t} * G_{k p}^{t} f_{p}\left(e_{p u b}^{k}(z_{pub})\right)+b_{t}^{pub})\\
\label{eq:taylor}
\end{equation}
\begin{equation}
\widehat{y_{t}^{pri}} =\sigma( \sum_{k^{\prime}=1}^{K_{p r l}} W_{k^{\prime}}^{t} * e_{p r i}^{k^{\prime}}(z_{pri})+b_{t}^{pri})
\label{eq:taylor1}
\end{equation}
\begin{equation}
\widehat{y_{t}} = \widehat{y_{t}^{pub}} + \widehat{y_{t}^{pri}}
\label{eq:taylor2}
\end{equation}
where $e_{pub}^k(\cdot)$ represents the $k$-th sharing DNN, $e_{pub}^{k^{\prime}}(\cdot)$ represents the $k^{\prime}$-th private expert layer, $f_p(\cdot)$ represents the $p$-th explicit mapping, $G_{kp}^t$ is the gated value corresponding to the $k$-th expert after its $p$-th explicit mapping for the $t$-th task, $W_{k p}^{t}$ and $W_{k^{\prime}}^{t}$ are parameters of task tower $\phi_t(\cdot)$, that is, the weights corresponding to each expert output, $\widehat{y_{t}}$ represents the predicion values of the $t$-th task.

\subsection{Virtual gradient Setting}\label{3.3}
Gating unit $G_{kp}^t$ in DEPHN is designed to properly control the information flow among shared experts and cater to different tasks. However, under the basic setting of MMOE, the gradient received by gating value is only related to the corresponding task, and the relationship between tasks cannot be obtained. For parameter $G_{kp}^t$, gradient received by it can be obtained by derivation of the chain rule:
\begin{equation}
    \frac{\partial L_{t}}{\partial G_{k p}^{t}}=\frac{\partial L_{t 1}}{\partial \hat{y}} * \frac{\partial \hat{y}}{\partial \sigma} * \frac{\partial \sigma}{\partial G_{kp}^t}=\left(\hat{y}_{t}-I(a)\right) * W_{kp}^t f_{p}(e_{pub}(z_{pub})).
\label{eq:chain}
\end{equation}
It can be seen that the gradient received by the gating parameter $G_{kp}^t$ in the framework of MMoE is only related to task $t$. Therefore, in the training process of DEPHN, we introduced a "virtual gradient coefficient" for the gated unit on the premise of not affecting the forward propagation value. This coefficient can be controlled by the label correlation between tasks and the difference of gated values. Specifically, the "virtual gradient coefficient" makes reasonable adjustments to gating through the correlation measure and difference measure of task $\mathcal{T}_t$ and task $\mathcal{T}_j$, that is, when a single gating unit receives the gradient of the corresponding task loss return, it also receives adjustments from other t-1 tasks.
\begin{equation}
    \frac{\partial L_{t}}{\partial G_{k p}^{t}}=\left(\hat{y}_{t}-I(a)\right) * W_{kp}^t f_{p}(e_{pub}(z_{pub}))*\prod_{j\ne t}^{}\gamma _{jt}.
\label{eq:chain2}
\end{equation}

It is reasonable that the same explicit mapping of the same expert can give similar information flows when task correlation is high, and give different information flows when task correlation is low. Therefore, we can use the differences of gating values to describe the difference of information flows. It is our purpose that the "virtual gradient coefficient" amplifies the gradient of the updated gated value when gating value difference and task correlation is both large(high) or small(low), and narrows the same gradient when gating value difference is small and task correlation is high or gating value difference is large and task correlation is low. The cosine function can be preliminarily used to represent the definition of this coefficient:
\begin{equation}
    \gamma_{\mathrm{jt}}=\cos \left(\left(\phi\left(Y_{j}, Y_{t}\right)+\min \left(1,\left|G_{k p}^{t}-G_{k p}^{j}\right|\right)\right) * \pi\right)+1,
\label{eq:scaling}
\end{equation}
where $\phi(Y_j,Y_t)$ represents the similarity of task labels within a batch (in implementation, the absolute value of the cosine similarity or Pearson correlation coefficient is used), $\left | G_{kp}^t-G_{kp}^j \right | $ is the difference of gating values for task $t$ and task $j$ by the same mapping of the same expert. By the property of cosine function, we can artificially control the gradient scaling in different cases. We discuss eight different types of function to measure gating value difference and record corresponding experimental results in section\ref{5}.

Since the gradient in the above backpropagation process is artificially designed, and we do not want to interfere with the normal activation value in the feedforward process. In the implementation, we will first calculate the corresponding difference scaling coefficient of the gated activation value, multiply it by the corresponding gating unit, and then divide it by its freezing term, so as to complete the setting of a "virtual gradient" in the back propagation.

\section{EXPERIMENTS}\label{4}
In this section, we evaluate DEPHN on two manual datasets based on baseline datasets and two real competition datasets. We aim to answer the following research questions:

\begin{itemize}
	\item {\bf RQ1:} Can part of the shared expert analysis layer constructed by different cross modes and corresponding design give better generalization effect by expanding the hypothesis space in the multi-task scenario?
	\item {\bf RQ2:} Whether explicit mapping structures and virtual gradients can provide enhancement for multi-task data set fitting. Which design function is more interpretable?
	\item {\bf RQ3:} For the real contest data set, whether DEPHN has better robustness and whether the explicit relationship it gives can better explain the correlation between tasks.
\end{itemize}

\subsection{Experimental Setting.}\label{4.1}
In order to evaluate the effectiveness of the DEPHN, the experiments select a benchmark click-through rate({\bf CTR}) data set  and two large multi-task competition data sets:

\begin{itemize}
	\item {\bf Criteo\footnote{https://www.kaggle.com/c/criteo-display-ad-challenge}:}
    Criteo is a benchmark dataset of CTR predictive tasks, including 46840617 samples and 1 task. Under the MTL objective, we use functional mapping of the base confidence to make multiple task objectives. The basic confidence of the dataset is obtained by a simple feed forward neural network to ensure the "fitability" of the information. Then, the relations $y_r=\frac{c}{abs(c)}*2ln(|c|+1)+0.1e^{|c|}+\epsilon$ and "$y_{ur}=\frac{c}{abs(c)}(2sin(|c|)+0.1cos(|c|))$" are used to determine the relevant task label (correlation coefficient, $\sigma=0.9877$) and the irrelevant task label (correlation coefficient, $sigma=0.5288$). Under the constant input characteristics, Criteo-R represents the relevant data set composed of basic confidence labels and related labels, while Criteo-UR represents the unrelated data set composed of basic confidence labels and unrelated labels. All the labels are determined by the indicative function after Determine the confidence.
	\item {\bf Ali-CCP:\footnote{https://tianchi.aliyun.com/dataset/408}}
    In the existing recommendation system multitasking model, researchers usually use Ali-CCP data set as a benchmark to verify the robustness of the model. It includes 85316519 samples and 2 tasks of CTR prediction and CVR prediction. The click label represents the event that the users click on item, while the conversion label represents the conversion time that occurs after the users click. Therefore, task dependence exists in Ali-CCP data set.
	\item {\bf WeChat-2021\footnote{https://algo.weixin.qq.com/}:}
    WeChat2021 is the training data set used in the preliminary competition of WeChat Big Data Challenge in 2021. It mainly covers some behaviors of users in short video scenes, including 7317882 samples and 7 tasks. In this paper, "\textit{read comment, like and click avatar}" are selected to verify the robustness of the model when more than two task objectives are pursued
\end{itemize}

In order to verify the effectiveness of the proposed model, we used the traditional CTR prediction model (DNN, W\&D\cite{widedeep}, DCN\cite{dcn}, FINT\cite{FINT}, NFM\cite{nfm}, xDeepFM\cite{xdeepfm}), and some benchmark multi-task models (MMoE\cite{MMOE}, PLE\cite{2020Progressive}, SNR\cite{2019SNR}) for experiment and verification. In order to keep the consistency of results in the experiment process, the hyper-parameters (such as the dimension of embedding, number of interaction layers, number of experts, learning rate, etc.) for the determination of model structure and training process are set uniformly.

\subsection{Parallel experts experiment (RQ1)}\label{4.2}
For research question 1, we first conducted the parallel structure design based on MMoE and SSG (the MTPHN structure), and independently verified the validity of its structures on real data sets.

\newcommand{\tabincell}[2]{\begin{tabular}{@{}#1@{}}#2\end{tabular}}
\begin{table}[ht]
  \centering
  \small
  \fontsize{9}{12}\selectfont
  \caption{Experimental results of MTPHN on Ali-CCP dataset}\label{table:exp1}
  \resizebox{0.48\textwidth}{!}{
    \begin{tabular}{c|c|cc|cc}
    \hline
    \multirow{2}*{Model} & \multirow{2}*{Expert setting} & \multicolumn{2}{c|}{Task 1} & \multicolumn{2}{c}{Task 2} \\
    \cline{3-6}
       & & logloss & AUC & logloss & AUC \\
    \hline
     DNN    & - & 0.161467 & 0.627484 & 0.003462 & 0.531357 \\
     W\&D   & - & 0.161684 & 0.624813 & {\bf 0.002457} & {\bf 0.618657} \\
     DCN    & - & 0.161539 & 0.626710 & 0.003462 & 0.528997 \\
     FINT   & - & {\bf 0.161060} & {\bf 0.630317} & 0.002604 & 0.610606 \\
    \hline
    \hline
     MMoE   & - & 0.161248 & 0.620267 & 0.002050 & 0.638755 \\
     PLE    & - & 0.161167 & 0.619016 & 0.002091 & 0.598048 \\
     SNR    & - & 0.161060 & 0.624165 & 0.002104 & 0.542347 \\
    \hline
     \multirow{3}*{MT-PHN} & $D_{pub}, C\&F_{pub}$ & {\bf 0.161049} & 0.622513 & {\bf 0.002021} & 0.637854 \\
                           & $D_{pri}, C\&F_{pub}$ & 0.161870 & 0.624355 & 0.002062 & 0.649029 \\
                           & $D_{pub}, C\&F_{pri}$ & 0.161390 & {\bf 0.625701} & 0.002056 & {\bf 0.655160} \\
    \hline
    \end{tabular}}
\end{table}

Table\ref{table:exp1} shows the validation experiments by different models on Ali-CCP data set. Result shows that, in the case of single task, the CTR prediction models have good stability, and on the CTR prediction task without dependencies, these models can usually give better results than the multi-task model. However, in the CVR prediction task, the single-task model could not give good results. On the one hand, the positive sample ratio of CVR was very low, which causes more serious sample bias during training. On the other hands, the auxiliary prediction information could not be obtained from the CTR task. In the multi-task model experiment, the model makes full use of the labels of different tasks to provide sufficient information guidance for model training. In the benchmark model, the MMoE structure better reflects the "multi-expert" performance, giving the best performance on the combined tasks of CTR and CVR.

Table\ref{table:exp1} also gives the corresponding experimental results for the sharing of different kinds of cross-analysis modules involved in MTPHN. The experimental results show that the explicit interaction method composed of Cross and Field interaction methods is more suitable for exploring the feature distribution cross mode of single object, while the nonlinear information fitting method composed of DNN is more suitable for mining the basic information shared between different tasks. Therefore, in the subsequent DEPHN, we determined the underlying analysis method that consists of DNN as a public module and Cross and Field Interaction as a private module.

\begin{figure}
  \centering
  \addtocounter{subfigure}{0}\subfigure[the heat map of parameter value in SSG]{
    \includegraphics[width=.21\textwidth]{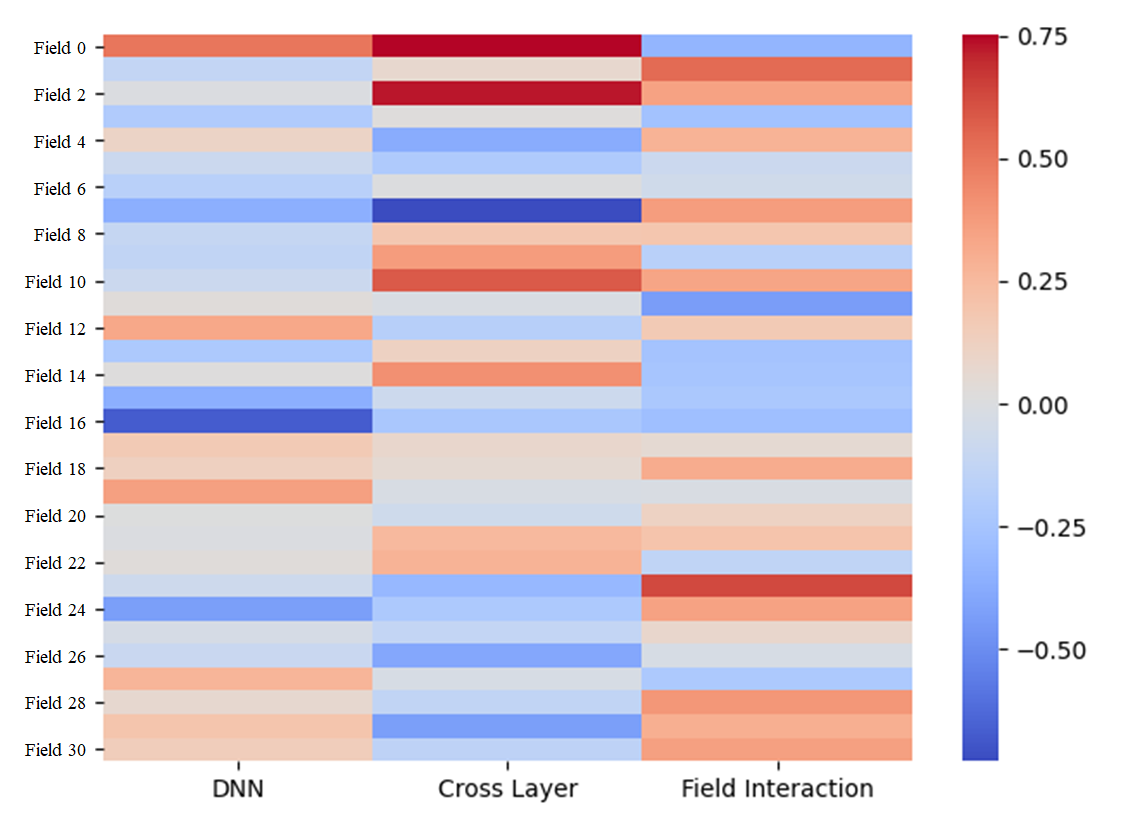}
  }
  \addtocounter{subfigure}{0}\subfigure[the ratio of public-private partial activation value]{
    \includegraphics[width=.21\textwidth]{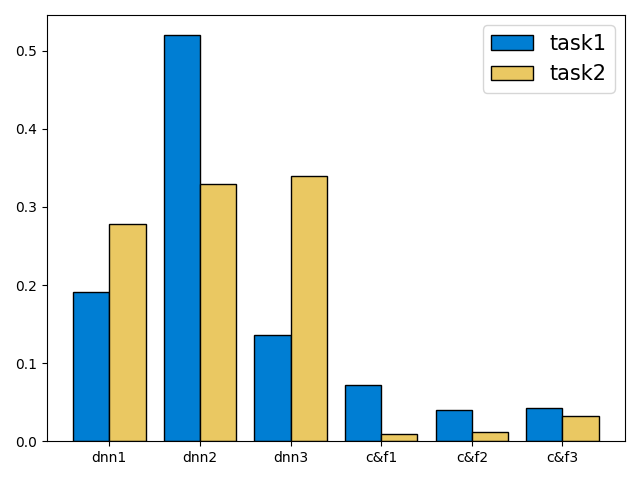}
  }
  \caption{{\bf The visualization of MTPHN performance in SSG and activation value}}
  \label{pic:mtphn}
\end{figure}

Figure\ref{pic:mtphn} is the heat map of parameter value in SSG and the ratio of public-private partial activation value. Figure\ref{pic:mtphn}(a) shows the heat map of soft gating parameters in the three SSG mapping results. The experimental results show that when the MSA enhancement feature and the raw feature are shared, the information flow components fitted by soft gating parameters for the three cross modes are different, indicating that the required feature information for different information cross modes is different. Figure\ref{pic:mtphn}(b) shows the contribution ratio of absolute activation value before the final activation function between public and private modules in different tasks. The experimental results show that, in the test set, most of the activation values given by the public modules, while the private modules supplement the information of a single task. Therefore, based on MTPHN, the research direction is mainly focused on "how to improve the information integration and expression ability of common modules for different tasks".

\subsection{Explicit mapping virtual gradient experiment(RQ2)}\label{4.3}

\begin{figure}[h]
\setlength{}
  \centering
  \addtocounter{subfigure}{0}\subfigure[$cos(2\pi xy)+1$]{
    \includegraphics[width=.10\textwidth]{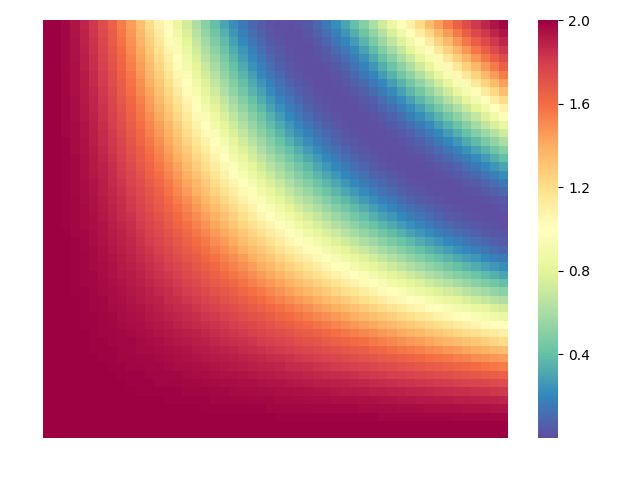}
  }
  \addtocounter{subfigure}{0}\subfigure[$2*|2xy-1|$]{
    \includegraphics[width=.10\textwidth]{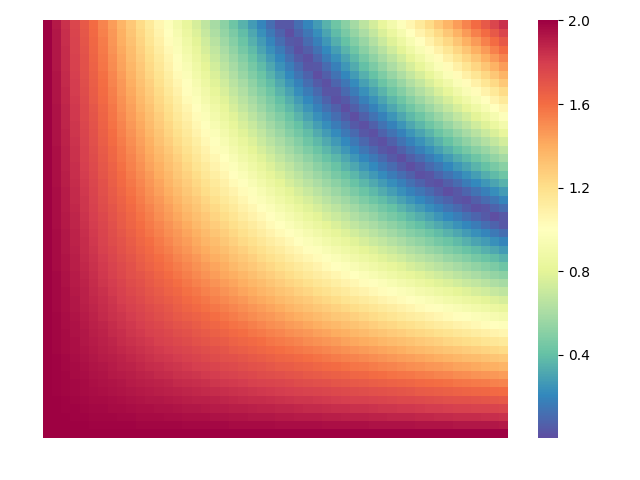}
  }
  \addtocounter{subfigure}{0}\subfigure[$2*(2xy-1)^2$ ]{
    \includegraphics[width=.10\textwidth]{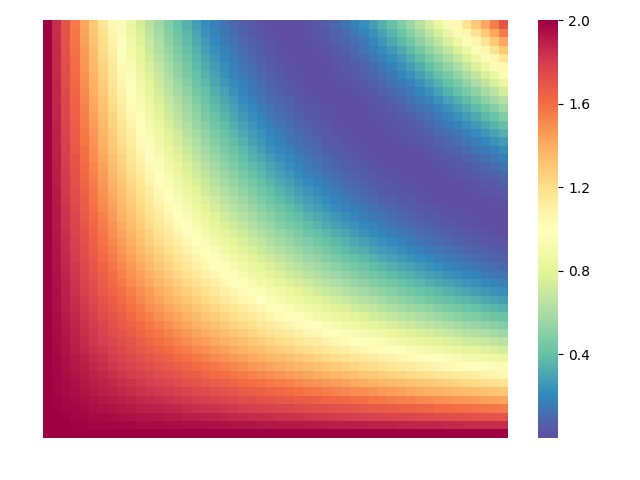}
  }
  \addtocounter{subfigure}{0}\subfigure[$2*(2xy-1)^{0.5}$]{
    \includegraphics[width=.10\textwidth]{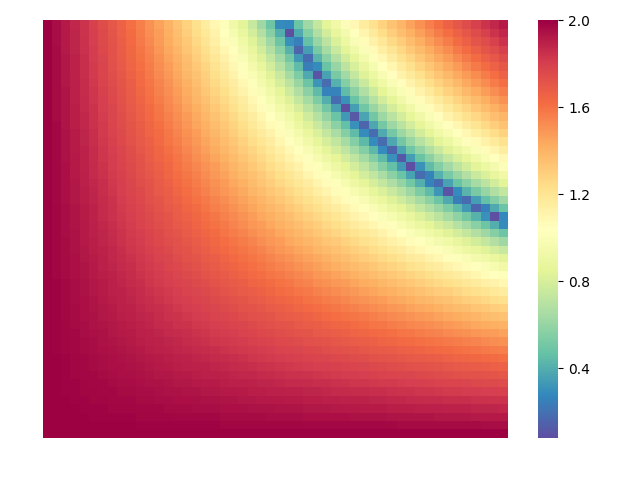}
  }
  \addtocounter{subfigure}{0}\subfigure[$cos(\pi x+\pi y)+1$]{
    \includegraphics[width=.10\textwidth]{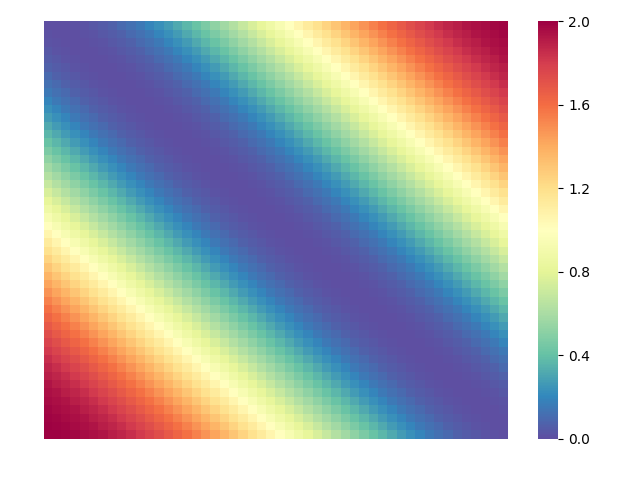}
  }
  \addtocounter{subfigure}{0}\subfigure[$2*|x+y-1|$]{
    \includegraphics[width=.10\textwidth]{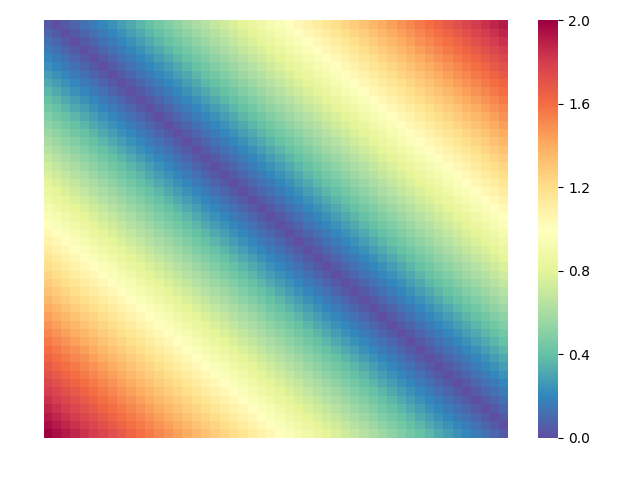}
  }
  \addtocounter{subfigure}{0}\subfigure[$2*(x+y-1)^2$]{
    \includegraphics[width=.10\textwidth]{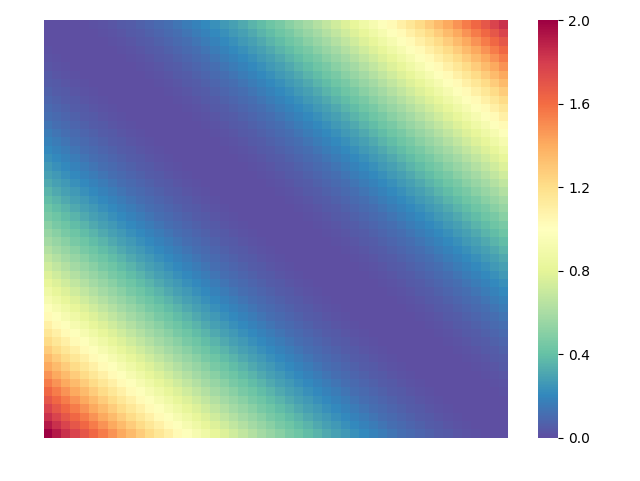}
  }
  \addtocounter{subfigure}{0}\subfigure[$2*(x+y-1)^{0.5}$]{
    \includegraphics[width=.10\textwidth]{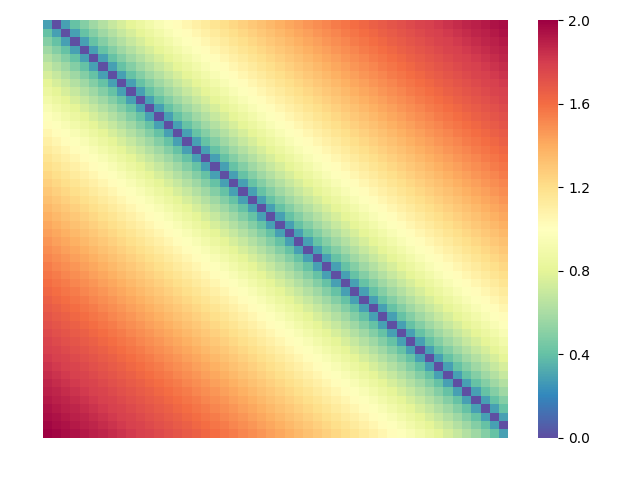}
  }
  \caption{The heat maps of different virtual gradient coefficient calculation function. In the figure, x represents the task relevance measure (Cosine similarity or Pearson correlation coefficient), y represents the gating variance. The values of x and y are both between 0 and 1.}
  \label{pic:heat-vg}
\end{figure}

In this section, we mainly use the manual data set based on Criteo to further verify the validity of DEPHN on the basis of MTPHN. Firstly, based on the requirements of the virtual gradient coefficient, we specify 8 different types of functions according to the interactive information mode and the fitness function. They determine the different virtual gradient parameters by determining the balance of the difference between the task correlation measure and the gated value with the setting function. Figure\ref{pic:heat-vg} shows the expression of 8 virtual gradient functions and gradient coefficient distribution heat maps.

The colors in the heat maps represent the positive and negative influences on the gradient, in which blue represents the reduced gradient, red represents the enhanced gradient, and the yellow region is the dividing line between the two influences. The figure\ref{pic:heat-vg} shows that, based on the interactive information of "multiplicity" (Subfigures (a), (b), (c), and (d)), the gradient influence value is concentrated in the upper right balance region of the difference between task relevance measure and gated value, while based on the interactive information mode of "additive" (Subfigures (e), (f), (g), and (h)), the gradient influence value is concentrated in the diagonal region to reduce the gradient.

\begin{figure}[h]
  \centering
  \addtocounter{subfigure}{0}\subfigure[Criteo-R with cosine similarity]{
    \includegraphics[width=.22\textwidth]{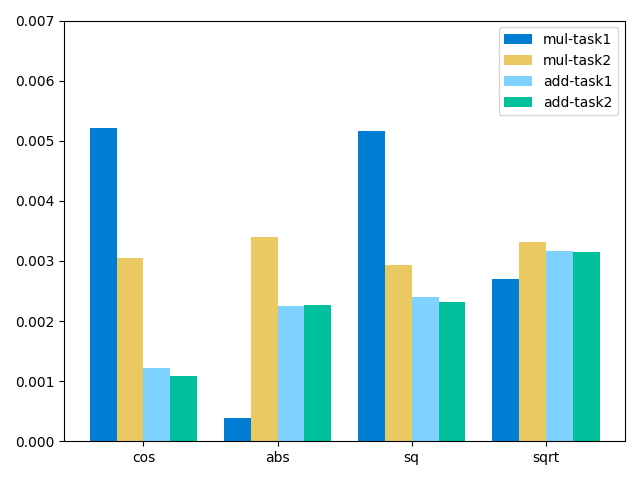}
  }\hspace{.01\textwidth}
  \addtocounter{subfigure}{0}\subfigure[Criteo-UR with cosine similarity]{
    \includegraphics[width=.22\textwidth]{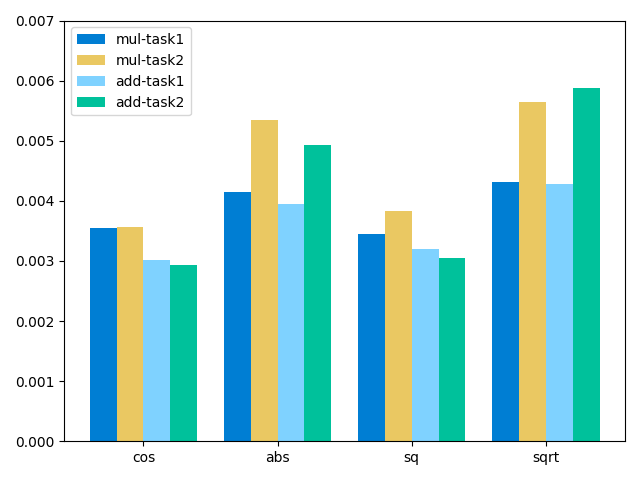}
  }
  \addtocounter{subfigure}{0}\subfigure[Criteo-R with Pearson correlation]{
    \includegraphics[width=.22\textwidth]{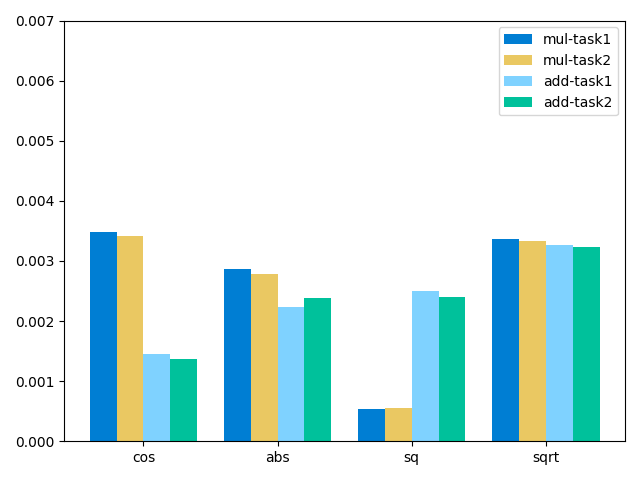}
  }\hspace{.01\textwidth}
  \addtocounter{subfigure}{0}\subfigure[Criteo-UR with Pearson correlation]{
    \includegraphics[width=.22\textwidth]{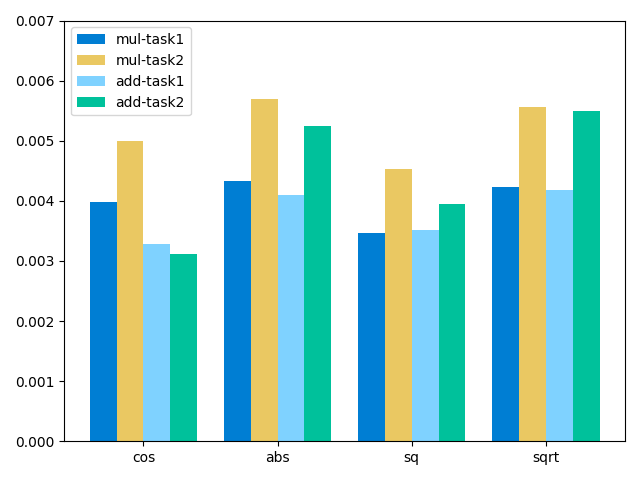}
  }
  \caption{Visualization of enhanced performance of different correlation measures on artificial data sets based on MMoE.}
  \label{pic:bar-vg}
\end{figure}

From the perspective of fitness function, the absolute value function basically meets the function setting requirements, cos function can reasonably magnify the influence value of the gradient coefficient, while the square function and the square root function have the function of decreasing and enlarging the gradient coefficient in a biased way. Figure\ref{pic:bar-vg} show the experimental results of DEPHN based on different correlation measures and virtual gradient functions on artificial data sets.

The experimental results show that, the representation ability of the model is generally improved by introducing the virtual gradient coefficient, and most of the functions show better results. The figure\ref{pic:bar-vg} shows the comparison effect of additive function and component function in the case of different correlation measures. It shows that, in the relevant data set, experimental results based on cosine similarity show a more obvious skew in the relevant data set, and the combination of Pearson correlation coefficient and additive square root function can often give better stability of multi-task results. On the uncorrelated data sets, the experimental results of the two kinds of correlation measures are basically the same, and the results given by the square root function are usually the most stable. Due to the low correlation between tasks in real life, DEPHN finally decided to use the combination of Pearson correlation coefficient and additive square root function as the method to calculate the virtual gradient coefficient based on the experimental results.

\begin{figure}[h]
  \centering
  \includegraphics[width=0.35\textwidth]{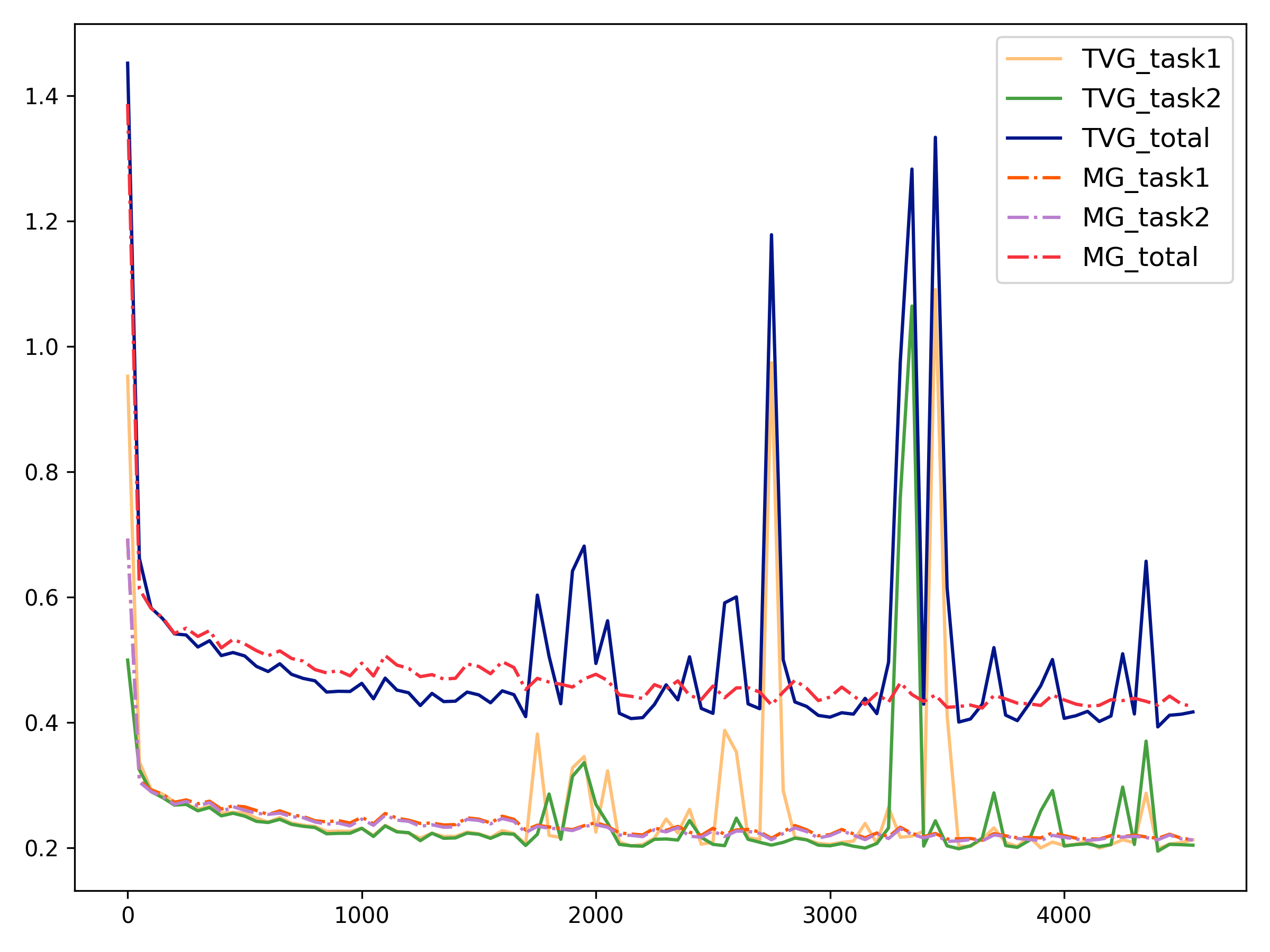}
  \caption{loss comparison diagram during DEPHN training under different gated units. TVG represents trainable value gating and MG represents mapping gating.}\label{pic:gating_compere}
\end{figure}

As a basic multi-task model, the gate value of MMoE is mapping gating, which is obtained based on the linear mapping of input features. It means that the gate value is a variable determined by the distribution of input features. In DEPHN, we tried to use gated differences to measure the difference in information flow differentiating from shared formation. From a global perspective, the gating value of variable version may not be able to give a more stable correction of gradient, so trainable fixed parameters were used in the model as the gating value.

The figure\ref{pic:gating_compere} shows the loss changes of parametric gating and mapping gating. The test results show that when mapping gating is used in DEPHN, the overall loss convergence tends to be smooth, and the late fluctuation is small, but the overall effect is not as good as global parameter gating, and it is easy to fall into the local optimal solution. Global parameter gating will explore the information differentiation mode of the model at the later stage of training, and reasonably improve the model performance through label correlation. From the changes of loss in different tasks, it can be seen that DEPHN based on parametric gating links the changes of task optimization together, thus leading more conducive to multi-objective optimization. This gives the model more exploration space and is conducive to finding the global optimal solution of the model parameters.

\begin{figure}[h]
\setlength{}
  \centering
  \addtocounter{subfigure}{0}\subfigure[Confidence of MMoE]{
    \includegraphics[width=.2\textwidth]{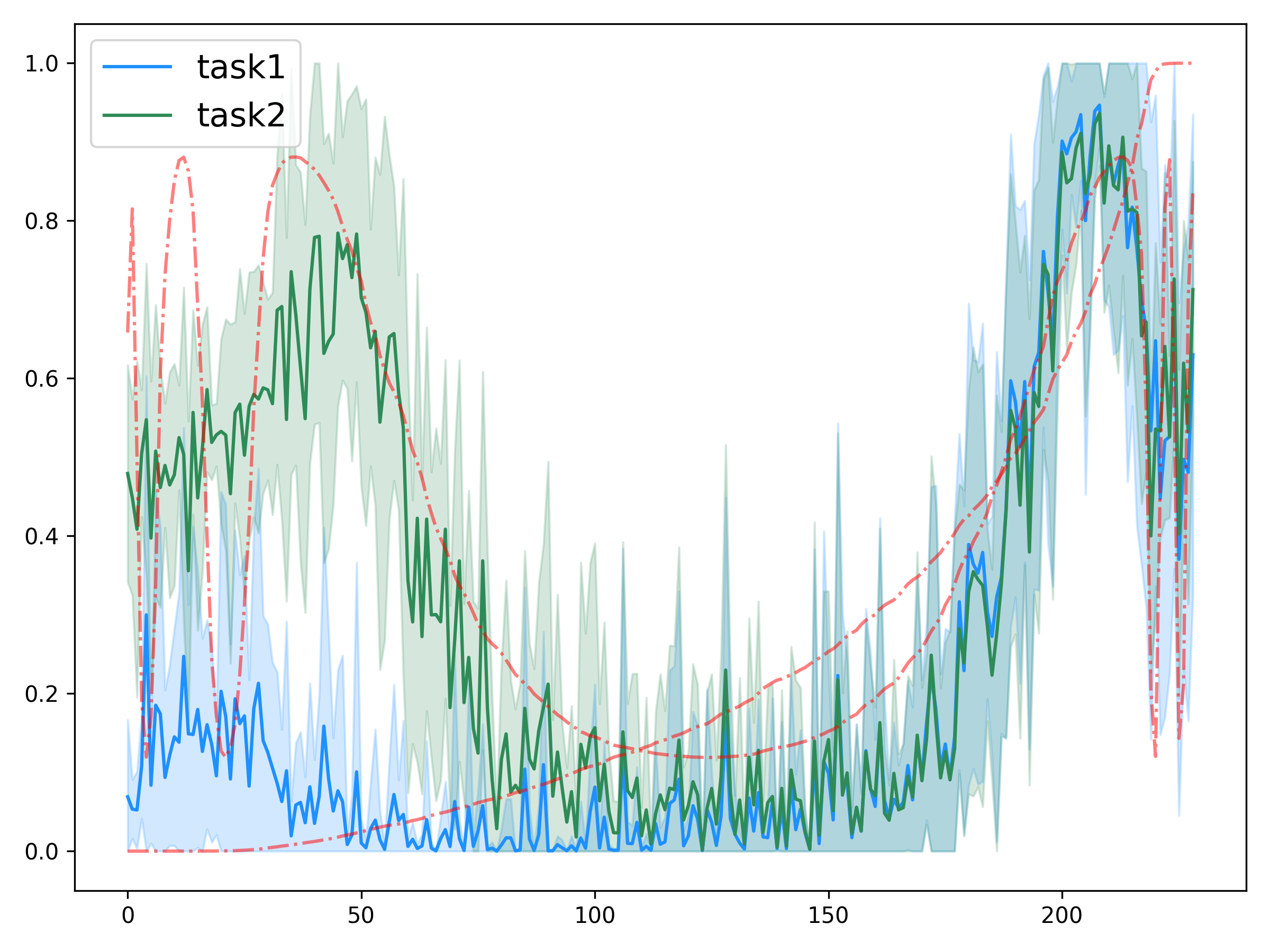}
  }
  \addtocounter{subfigure}{0}\subfigure[Confidence of DEPHN]{
    \includegraphics[width=.2\textwidth]{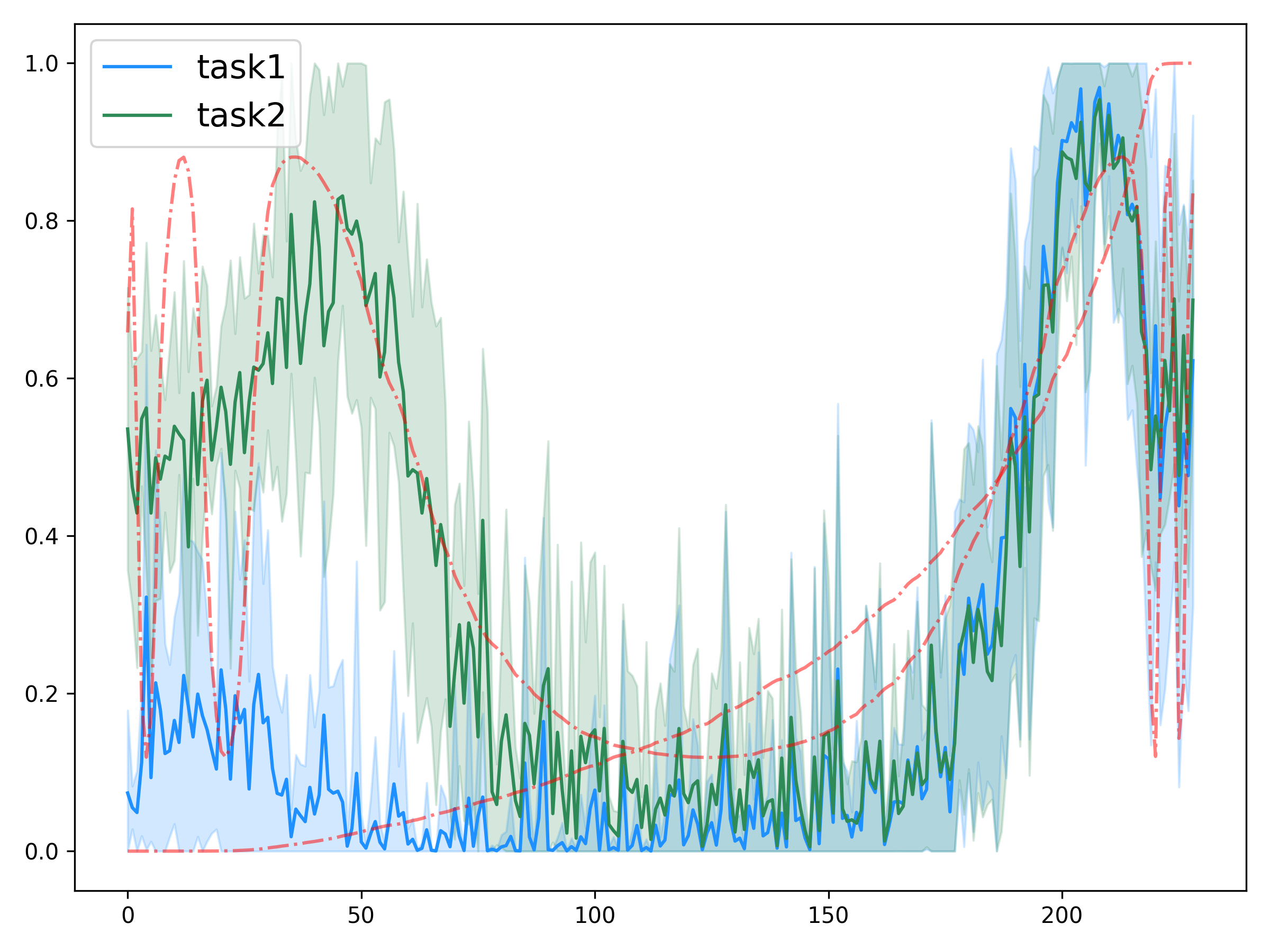}
  }
  \caption{The activation value of MMoE and DEPHN on Criteo-UR. The red line represents the basic potential confidence level set by the artificial data set for task 1 and Task 2 respectively, the blue area represents the prediction interval for Task 1, and the green area represents the green interval for task 2}
  \label{pic:active}
\end{figure}

Figure\ref{pic:active} shows the hidden activation values of the MMoE model and DEPHN model for unrelated data sets respectively. The result shows that, compared with MMoE, DEPHN can better arrange the relationship between activation values in the region with smooth relationship, and better separate the confidence of positive and negative samples when the activation values are near both ends.

\subsection{Performance experiment (RQ3)}\label{4.4}

\begin{table*}[ht]
  \centering
  \small
  \fontsize{9}{12}\selectfont
  \caption{Performance results of different models on large competition data sets}\label{table:exp3}
  \resizebox{\textwidth}{!}{
    \begin{tabular}{c|c|cc|cc|cc|cc|cc}
    \hline
    \multirow{3}*{Target} & \multirow{3}*{Model} & \multicolumn{4}{c|}{Ali-CCP} & \multicolumn{6}{c}{WeChat-CP2021} \\
    \cline{3-12}
       & & \multicolumn{2}{c|}{Task 1} & \multicolumn{2}{c|}{Task 2} & \multicolumn{2}{c|}{Task 1} & \multicolumn{2}{c}{Task 2} & \multicolumn{2}{c}{Task 3} \\
    \cline{3-12}
       & & logloss & AUC & logloss & AUC & logloss & AUC & logloss & AUC & logloss & AUC\\
    \hline
    \multirow{5}*{\tabincell{c}{Single\\Task\\Learning}}
        & DNN     & 0.161467 & 0.627484 & 0.003462 & 0.531357 & 0.095421 & 0.925291 & 0.091059 & 0.826466 & 0.037159 & 0.832976 \\
        & W\&D    & 0.161684 & 0.624813 & 0.002457 & {\bf 0.618657} & 0.093197 & {\bf 0.926729} & 0.090587 & 0.827845 & 0.037363 & 0.831841 \\
        & DCN     & 0.161539 & 0.626710 & 0.003462 & 0.528997 & 0.094495 & 0.924850 & 0.090668 & 0.828572 & 0.037012 & 0.832986 \\
        & FINT    & {\bf 0.161060} & {\bf 0.630317} & 0.002604 & 0.610606 & 0.093764 & 0.925961 & 0.090373 & 0.830775 & {\bf 0.036682} & 0.839455 \\
        & NFM     & 0.161971 & 0.617122 & {\bf 0.002310} & 0.589522 & {\bf 0.096288} & 0.917460 & 0.092024 & 0.818766 & 0.037966 & 0.814771 \\
        & xDeepFM & 0.170470 & 0.602327 & 0.004157 & 0.558556 & 0.093896 & 0.925278 & {\bf 0.090316} & {\bf 0.832673} & 0.036780 & {\bf 0.840571} \\
    \hline
    \hline
    \multirow{4}*{\tabincell{c}{Multi\\Task\\Learning}}
        & MMoE    & 0.161248 & 0.620267 & 0.002050 & 0.638755 & {\bf 0.093932} & {\bf 0.924594} & 0.091325 & 0.822586 & 0.037745 & 0.825175 \\
        & PLE     & 0.161167 & 0.619016 & 0.002091 & 0.598048 & 0.095841 & 0.918362 & 0.091993 & 0.819099 & 0.037880 & 0.808978 \\
        & SNR     & {\bf 0.161060} & 0.624165 & 0.002104 & 0.542347 & 0.096817 & 0.913339 & 0.091150 & 0.825172 & 0.038583 & 0.803636 \\
    \cline{2-12}
        & MT-PHN  & 0.161390 & 0.625701 & 0.002056 & 0.655160 & 0.094612 & 0.923507 & 0.091312 & 0.826003 & 0.037036 & 0.828731 \\
        & DE-PHN  & 0.162253 & {\bf 0.625885} & {\bf 0.002045} & {\bf 0.663543} & 0.101615 & 0.923981 & {\bf 0.091039} & {\bf 0.828510} & {\bf 0.037001} & {\bf 0.829934} \\
    \hline
    \end{tabular}}
\end{table*}

Based on the discussion of the above experiment, this section mainly uses two real large contest data sets to discuss the robustness of DEPHN. The following table shows the performance results of classic CTR estimation model, multi-tasking model, MT-PHN and DE-PHN in Ali-CCP and Wecht-2021 data sets in the recommendation system. Table\ref{table:exp3} shows the performance of experiments.

From the experimental results, we can know that the performance of the single task model is usually relatively stable. In Ali-CCP data set, the single-task model usually has a good understanding of CTR tasks, but a poor performance of CVR correlation tasks. This indicates that joint training can usually help the model to improve its understanding of the data environment when there is a dependency between tasks. In Wechat-2021 data set, the single-task model shows consistent results for different task objectives, and the performance is better than that of the multi-task model, indicating that the dependency between tasks is low in wechat-2021 data set.

In the multi-task model test, MMoE is usually stable, while PLE and SNR show the unbalanced phenomenon of generalization model, which is usually caused by information flow disorder in the multi-task target. On two large real data sets, MT-PHN and DE-PHN can fit the data sets well and maintain good equilibrium stability. On Ali-CCP data set, DE-PHN gives the SOTA performance of task 2, fully demonstrating that under the scenario of multi-objective optimization of recommendation system, the design of DEPHN fully meets the two goals of "generalization and understanding improvement of underlying shared information" and "good differentiation of information flow of different tasks".

\section{Conclusion}\label{5}
In this paper, a model named DEPHN is proposed for multi-task optimization under the background of recommendation system. In terms of basic information sharing, DEPHN uses multiple feature interaction methods in CTR prediction model to form heterogeneous expert analysis layer, and considers shared information and explict information between different tasks by SSG module and private interaction methods in input feature and analysis part. In terms of shared information differentiation, DEPHN starts from shared information flow differentiation and uses functions of different paradigms to find potential activation value relationships between different tasks. In the training process, DEPHN introduced a virtual gradient coefficient into the gated unit to adjust the activation value by using the differences of task relevance and gated value on the premise of not affecting the transmission of the activation value, which caters to the setting of information distribution differentiation. The experimental results show that DEPHN has a good fitting effect on both artificial data sets and two large benchmark competition data sets, which improves the model's ability to find the global optimal solution and also provides a certain interpretability for the task information flow correlation.

\bibliographystyle{IEEEtran}
\bibliography{PHN-MTL}

\end{document}